\documentclass{article}

\usepackage{arxiv}

\usepackage[utf8]{inputenc} 
\usepackage[T1]{fontenc}    
\usepackage{hyperref}       
\usepackage{url}            
\usepackage{booktabs}       
\usepackage{amsfonts}       
\usepackage{nicefrac}       
\usepackage{microtype}      
\usepackage{amsmath}
\usepackage{cleveref}       
\usepackage{graphicx}
\usepackage[super,comma,numbers]{natbib}
\usepackage{doi}
\usepackage{wrapfig2}
\usepackage{caption}
\usepackage{float}

\title{ConvPE: Convolutional Position Embedding Improves Transformers}
\date{}
\author{Zehao Wang \\
	Nanjing University\\
	\texttt{wzhagi@163.com} \\
}

\begin{document}
\maketitle

\begin{abstract}

ConvPE (Convolutional Position Embedding) improves both the performance and efficiency of Transformer models in language modeling, long-context extension, and vision. It injects a windowed, content-dependent bias into attention scores through a lightweight causal convolution over a low-dimensional projection of the hidden states. As a lightweight plug-in, ConvPE is compatible with RoPE, attention masks, and KV concatenation. It also supports high-dimensional position embeddings via high-dimensional convolutions. In a reproduction, ConvPE outperforms five 1D baselines and eight 2D baselines. The code is publicly available at \url{https://github.com/wzhagi/convpe}.

\end{abstract}
\keywords{Position Embedding \and Language Models \and Transformers \and Attention \and Vision Transformers}

\section{Introduction}

\begin{wrapfigure}{r}{0.48\textwidth}
    \centering
    \includegraphics[width=\linewidth]{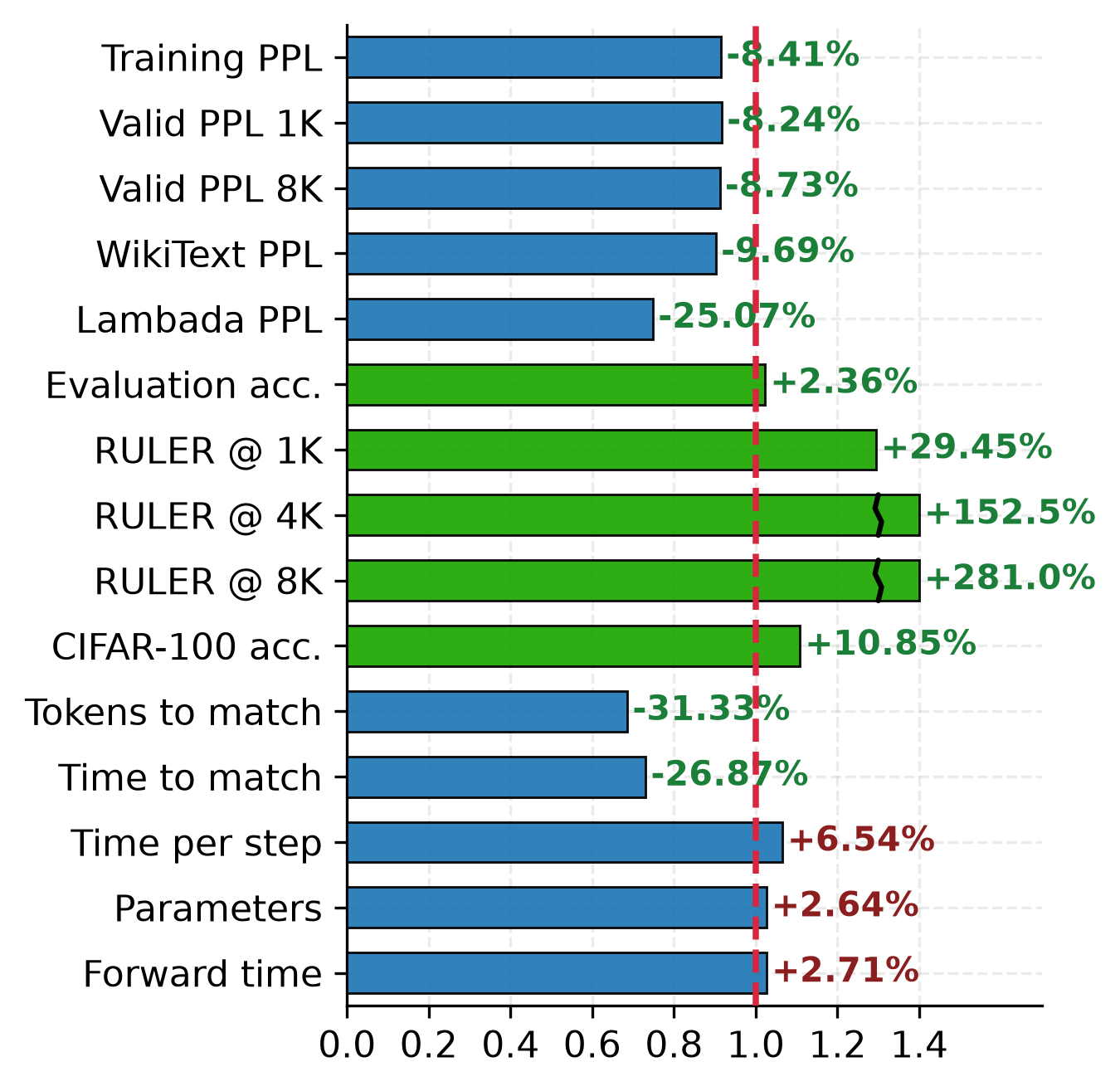}
    \captionsetup{hypcap=false}
    \caption{ConvPE relative to the RoPE baseline (dashed line at 1.0). Blue bars are lower-is-better metrics; green bars are higher-is-better metrics.}
    \label{fig:overview}
\end{wrapfigure}
\enlargethispage{3\baselineskip}

Transformers~\cite{vaswani2017attention} are insensitive to input order, whereas word order is crucial for language understanding and the spatial layout of image patches matters for vision. Explicit or implicit position embeddings are therefore needed to inject order information into the model~\cite{devlin2019bert,radford2019language,su2021roformer}. How this information is injected affects accuracy as the context length grows.

Position information is currently injected at one of three points. First, absolute position embeddings, whether fixed or learnable, add a position vector to token representations~\cite{vaswani2017attention,devlin2019bert,radford2019language}. Second, rotary position embedding (RoPE)~\cite{su2021roformer} rotates queries and keys so that attention scores depend on relative distance; RoPE has been widely adopted in modern large language models. Extending these methods to longer contexts requires interpolation or frequency rescaling~\cite{peng2024yarn}. Third, attention-score-bias methods add a term directly to the score matrix: ALiBi~\cite{press2021train} uses a fixed linear decay but performs worse on downstream tasks as the context length increases~\cite{kazemnejad2023impact}. CABLE~\cite{veisi2025cable} learns to modulate a fixed distance-penalty skeleton. The Composite method~\cite{chang2021convolutionsselfattentionreinterpretingrelative} adds convolution-based relative-position terms similar to ours but performs worse. DAPE~\cite{zheng2025dapev2} convolves the full score map, which is effective but incurs high computational overhead.

ConvPE follows this score-bias route and generates the bias locally. It applies a causal convolution to the hidden state projections and adds the resulting windowed bias directly to the attention scores. Beyond the window the bias is exactly zero, following NoPE methods~\cite{kazemnejad2023impact}, so distant keys are not penalized by construction. As a lightweight plug-in, ConvPE is compatible with RoPE, attention masks, and KV concatenation. It also supports high-dimensional position embedding through high-dimensional convolutions.

We evaluate ConvPE on language-model pretraining, downstream evaluation, long-context extension, and image classification. \Cref{fig:overview} summarizes the relative changes with respect to the RoPE baseline. Compared with RoPE, ConvPE achieves lower training and validation perplexity and higher benchmark accuracy, with minimal additional parameters and computational overhead. It reaches the baseline's final loss using fewer training tokens and less wall-clock time. After context-extension training to 8K tokens, the gap widens. As a plug-in, it also improves pretrained RoPE models during context-extension training. This demonstrates the potential of ConvPE for long-context language modeling. In vision, ConvPE2D improves CIFAR-100 accuracy over eight baselines.

\textbf{The main contributions of this paper are as follows:}
\begin{itemize}
    \item We propose ConvPE, a convolutional attention-score-bias position embedding that combines strong performance with broad compatibility.
    \item We study the influence of ConvPE hyperparameters and architecture choices, identifying a simple default configuration with a favorable performance-efficiency balance.
    \item We demonstrate that ConvPE improves training efficiency, token efficiency, and long-context extrapolation in language models.
    \item We verify the effectiveness of ConvPE2D in vision transformers on CIFAR-10 and CIFAR-100.
    \item We develop high-performance TileLang~\cite{DBLP:journals/corr/abs-2504-17577} kernels for ConvPE.
\end{itemize}

\section{Background}

\subsection{Attention and Score Biases}

Given the input query $Q \in \mathbb{R}^{S_q \times D_k}$, key $K \in \mathbb{R}^{S_{kv} \times D_k}$, and value $V \in \mathbb{R}^{S_{kv} \times D_v}$, the computation of scaled dot-product attention~\cite{vaswani2017attention} $O\in \mathbb{R}^{S_q \times D_v}$ is
\begin{equation}
O = \mathrm{Attention}(Q,K,V) = \mathrm{softmax}\left(\frac{QK^\top}{\sqrt{D_k}} + \mathbf{B} + \mathbf{M}\right)V,
\end{equation}
where $S_q$ is the query sequence length, $S_{kv}$ is the key/value sequence length, $\mathbf{B}\in\mathbb{R}^{S_q\times S_{kv}}$ is an additive attention-score bias, and $\mathbf{M}\in\mathbb{R}^{S_q\times S_{kv}}$ is a mask. A causal model sets $M_{ij}=-\infty$ for $j>i$, so that a query at position $s$ can attend only to keys at positions $t\le s$. With batch size and multi-head settings, $Q,K,V,O$ and the two bias terms carry additional batch and head dimensions; we omit them when they are unambiguous.

\subsection{Position Embeddings}

Existing position embedding schemes can be roughly divided into five categories:

\begin{enumerate}
    \item \textbf{Absolute Position Embedding}~\cite{vaswani2017attention}: adds a fixed positional vector, usually sine waves, to the input. Its performance is usually limited.
    \item \textbf{Learnable Positional Bias}~\cite{devlin2019bert,radford2019language}: adds a learnable positional vector to the input sequence. It is typically used when the maximum length is fixed, such as in fixed-resolution images or text truncated to a fixed length.
    \item \textbf{Rotary Position Embedding}~\cite{su2021roformer}: encodes position as a rotation matrix and applies it to queries and keys, so that attention scores depend on relative positions. RoPE has been widely adopted in modern large language models, and it enables long-context extension through frequency scaling, such as YaRN~\cite{peng2024yarn}.
    \item \textbf{Attention Score Bias}: adds a positional bias $\mathbf{B}$ to attention scores~\cite{raffel2020exploring,press2021train,chang2021convolutionsselfattentionreinterpretingrelative}. RoPE can be viewed as a rotation-based equivalent, as it also makes attention scores depend on relative positions.
    \item \textbf{No Position Embedding (NoPE)}~\cite{kazemnejad2023impact}: adds no explicit position embeddings; instead, it relies on the model structure and causal attention masks to learn positional information implicitly. For example, Kimi K3 applies NoPE to its global-attention layers~\cite{kimi2026k3}.
\end{enumerate}

\section{Method}
\label{sec:method}

\subsection{Windowed Convolutional Score Bias}

\begin{figure}[!ht]
    \centering
    \begin{minipage}{0.52\textwidth}
        \centering
        \includegraphics[width=\linewidth]{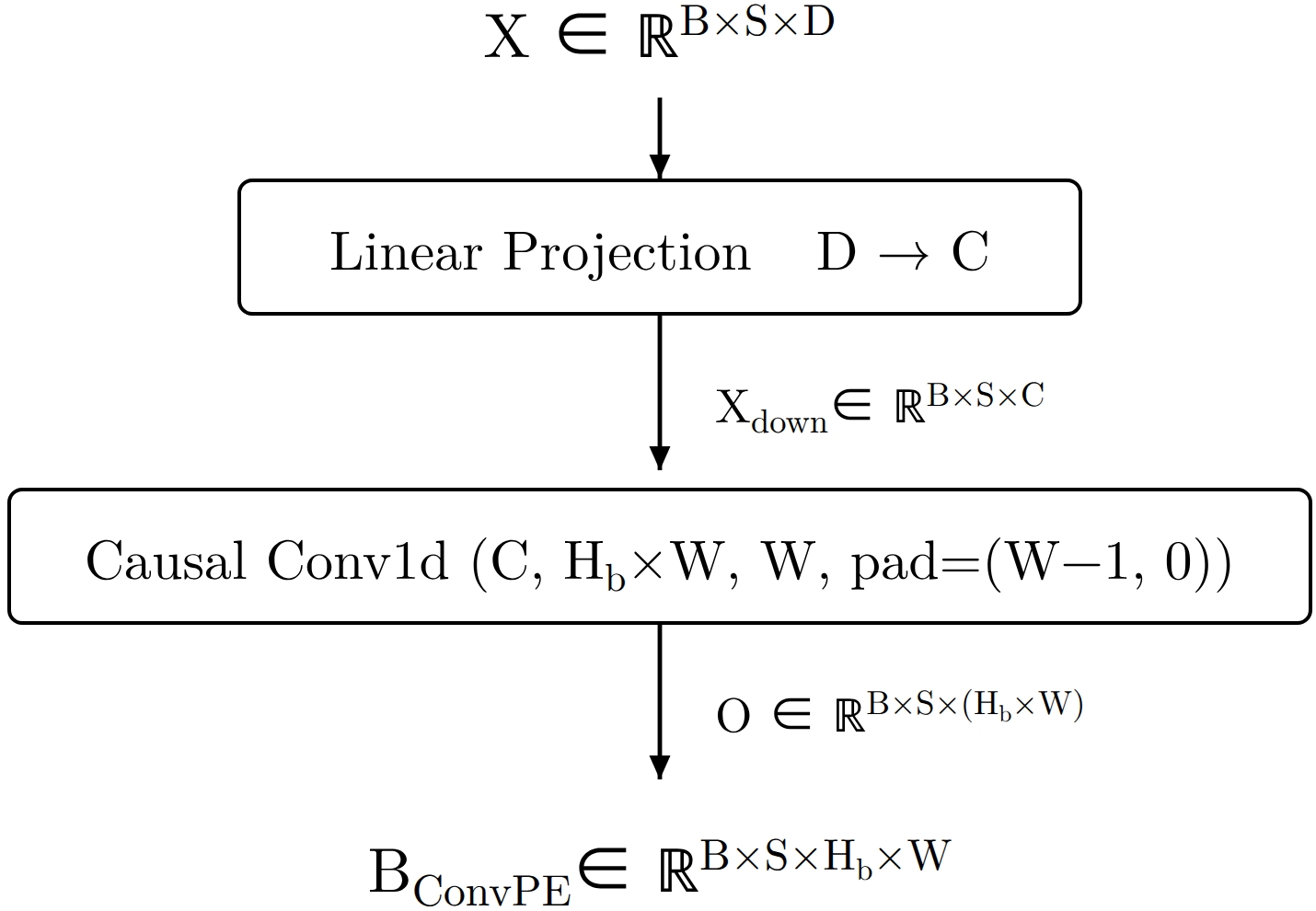}
    \end{minipage}
    \hfill
    \begin{minipage}{0.38\textwidth}
        \centering
        \includegraphics[width=\linewidth]{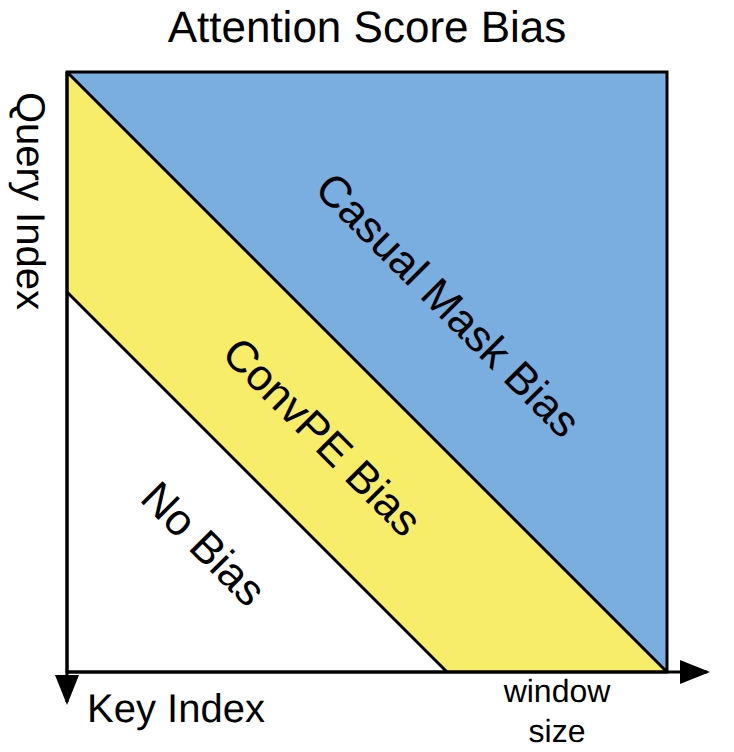}
    \end{minipage}
    \caption{Left: ConvPE algorithm. Right: attention score bias pattern.}
    \label{fig:convpe_method}
\end{figure}

Given token representations $X\in\mathbb{R}^{N\times S\times D}$, ConvPE first projects them to a small channel dimension,
\begin{equation}
X_{\mathrm{down}} = X W_p, \qquad W_p\in\mathbb{R}^{D\times C},
\end{equation}
and then applies a causal convolution with window $W$,
\begin{equation}
O_{\mathrm{Conv}} = \mathrm{CausalConv1d}\left(X_{\mathrm{down}};\, W,\, C\to H_bW\right),
\end{equation}
where $O_{\mathrm{Conv}}\in\mathbb{R}^{N\times S\times H_bW}$ and the convolution is left-padded by $W-1$ positions. The output is reshaped into the score bias
$\mathbf{B}_{\mathrm{ConvPE}}\in\mathbb{R}^{N\times S\times H_b\times W}$. For a query at position $s$ and a key at position $t\le s$, the bias added to the attention score of head $h$ is
\begin{equation}
\mathbf{B}_{n,h,s,t}=
\begin{cases}
\mathbf{B}^{\mathrm{ConvPE}}_{n,s,\,h\bmod H_b,\,s-t}, & 0\le s-t<W,\\[2pt]
0, & \text{otherwise},
\end{cases}
\end{equation}
where $N$ is the batch size, $H$ is the number of attention heads, and $H_b$ must divide $H$. In other words, head $h$ reads its bias from bias head $h\bmod H_b$, so each bias head is shared by $H/H_b$ attention heads. Each bias head therefore learns a different relative-position pattern. Future keys are removed by the causal mask, and keys older than $W$ are scored without a positional bias. \Cref{fig:convpe_method} illustrates the module and the resulting bias pattern.

The hyperparameters are:
\begin{itemize}
    \item $W$: convolution window size.
    \item $C$: output dimension of the linear projection and input channels of the convolution.
    \item $H_b$: number of score-bias heads.
\end{itemize}

The projection has $DC$ parameters and the convolution has $C H_b W^2$ parameters. Together with the $H_bW$ bias terms, a layer has $DC + C H_b W^2 + H_b W$ parameters. Each parameter is used once per token, so the multiply--accumulate count per token is the same.

\subsection{Parameter-Efficient Variants}

To reduce the convolution parameter count, we introduce three additional hyperparameters:
\begin{itemize}
    \item $\mathrm{Groups}$: number of convolution groups, which split the $C$ input channels and the $H_bW$ output channels into $G$ groups. Each group has its own kernel, and the parameter count becomes $C H_b W^2/G$.
    \item $\mathrm{Share}$: whether to share one kernel across all groups. Sharing divides the convolution parameter count by another factor of $G$ (for $G=1$ it has no effect).
    \item $\mathrm{Rank}$: rank $R$ of a low-rank factorization of the convolution. A group-shared down-projection maps the $(C/G)W$ input window to $R$ dimensions, and a group-wise up-projection maps $R$ back to the $H_bW$ outputs. This costs $(C/G)WR + R H_b W$ weight parameters instead of $C H_b W^2/G$, plus the same $H_bW$ bias term.
\end{itemize}
The default configuration uses $G=1$, no sharing, and $R=0$ (no factorization).

\section{Experiments}
\label{sec:experiments}

\subsection{Setup}
\label{sec:setup}

\textbf{Architecture.} Unless stated otherwise, all language models have $N_{\mathrm{layers}}=12$ and use pre-RMSNorm, SwiGLU~\cite{shazeer2020glu} feed-forward layers with $8D_{\mathrm{model}}^2$ parameters, multi-query attention (MQA)~\cite{shazeer2019fast} with a single key-value head ($K=V$)~\cite{deepseek2026v4}, query L2 normalization, key RMSNorm, a learnable attention sink per head, and tied embedding-layer and output-head weights. The tiny backbone has $D_{\mathrm{model}}=192$, $N_{\mathrm{heads}}=3$, and $D_{\mathrm{head}}=64$, and the base backbone has $D_{\mathrm{model}}=768$, $N_{\mathrm{heads}}=12$, and $D_{\mathrm{head}}=128$. The RoPE baseline uses a rotary dimension of $64$ and base $10000$. The vocabulary has $50{,}304$ entries (the GPT-2 tokenizer~\cite{radford2019language} padded to a multiple of 128).

\textbf{Data.} For hyperparameter search, we train on WikiText-103~\cite{merity2016pointer} and validate on the WikiText-2 test split. For pretraining, we use the first 3B tokens of FineWeb-Edu-10BT~\cite{penedo2024fineweb}. Context-extension training uses another 150M tokens. We hold out a disjoint 10M-token subset for validation.

\textbf{Optimization.} The language models are trained with the Hugging Face Trainer~\cite{wolf2020transformers}. The hyperparameter search uses AdamW~\cite{loshchilov2019decoupled} with a learning rate of $3\times10^{-4}$, weight decay $0.01$, a cosine learning-rate schedule, batch size $4$, sequence length $1024$, one epoch, BF16, gradient clipping at $1.0$, and seed $42$. For pretraining, Muon~\cite{jordan2024muon} optimizes the hidden 2D parameters with a learning rate of $3\times10^{-3}$, momentum $0.95$, and weight decay $0.01$, while Adam~\cite{DBLP:journals/corr/KingmaB14} optimizes the remaining parameters with a learning rate of $3\times10^{-4}$. Both use a cosine scheduler with 500 warm-up steps. We use a batch size of $4$ with $4$ gradient-accumulation steps, with sequence length $1024$, BF16 autocast, gradient clipping at $4.0$, and seed $42$. For context-extension training, we use the same settings as in pretraining except that we use $10\%$ of the pretraining learning rates and a batch size of $1$. All language-model results come from a single run; small task-level differences should be interpreted with sampling noise in mind.

\subsection{Hyperparameter Search}
\label{sec:hp}

We train 10 tiny and 8 base configurations and select by validation loss after one epoch. Model names record the backbone (\texttt{tiny} or \texttt{base}), the position mechanism (\texttt{rope} or \texttt{nope}), and the ConvPE fields $W$, $C$, $R$, $H_b$, and $G$ as \texttt{w}, \texttt{d}, \texttt{r}, \texttt{h}, and \texttt{g} followed by their values, with a trailing \texttt{s} when the convolution kernel is shared across groups. For example, \texttt{tiny rope w128d6r128h3g1s} denotes a tiny backbone with RoPE, $W=128$, $C=6$, $R=128$, $H_b=3$, $G=1$, and weight sharing. Fields equal to the default (position mechanism \texttt{nope}, $R=0$, $H_b=1$, $G=1$, and no sharing) are omitted when this is unambiguous, e.g., \texttt{base w128d16} is \texttt{base nope w128d16r0h1g1}.

\begin{figure}[!ht]
    \centering
    \begin{minipage}{0.48\textwidth}
        \centering
        \includegraphics[width=\linewidth]{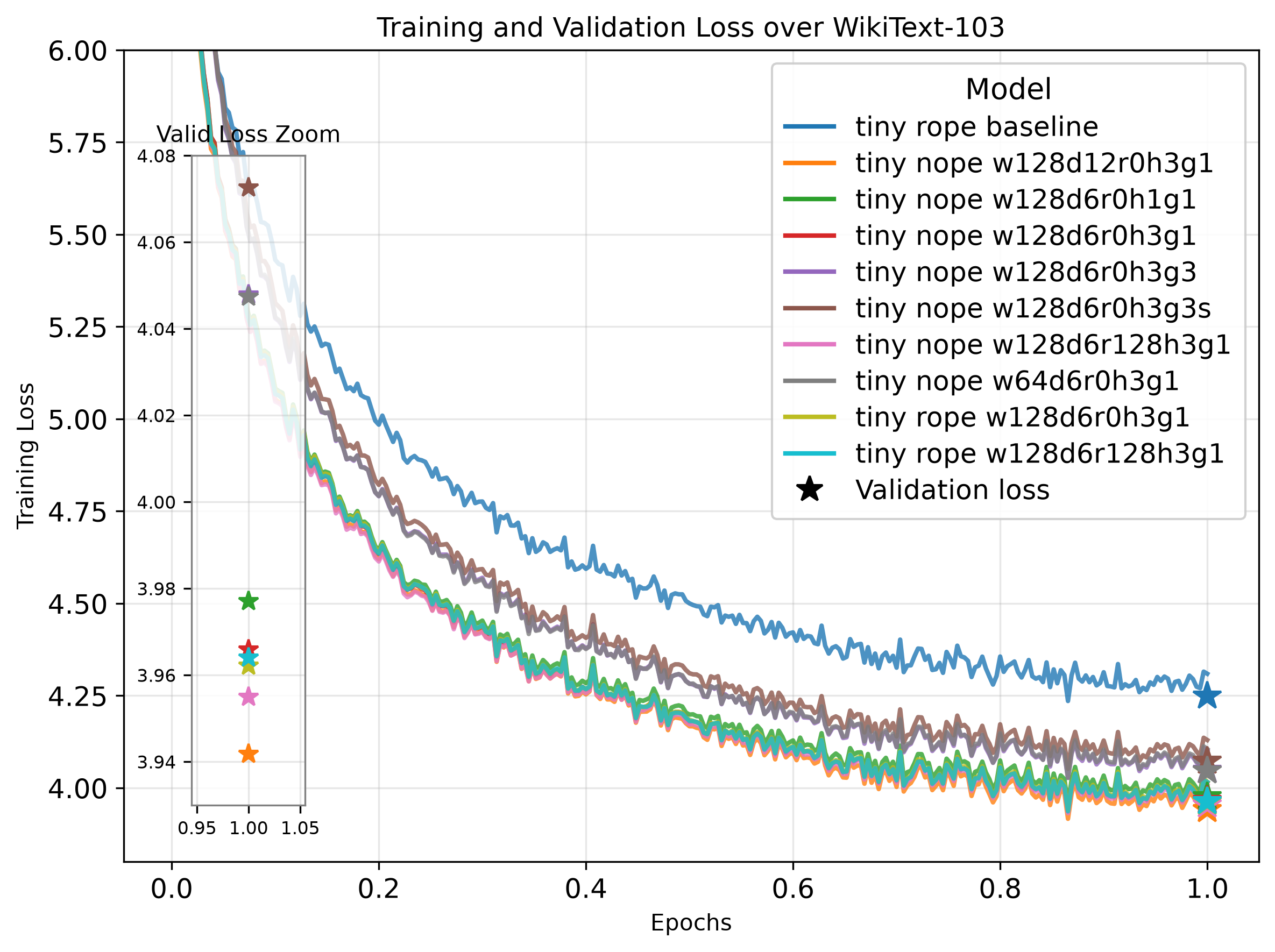}
        \caption{Training curves and validation loss of Tiny models on WikiText-103.}
        \label{fig:wt103t}
    \end{minipage}
    \hfill
    \begin{minipage}{0.48\textwidth}
        \centering
        \includegraphics[width=\linewidth]{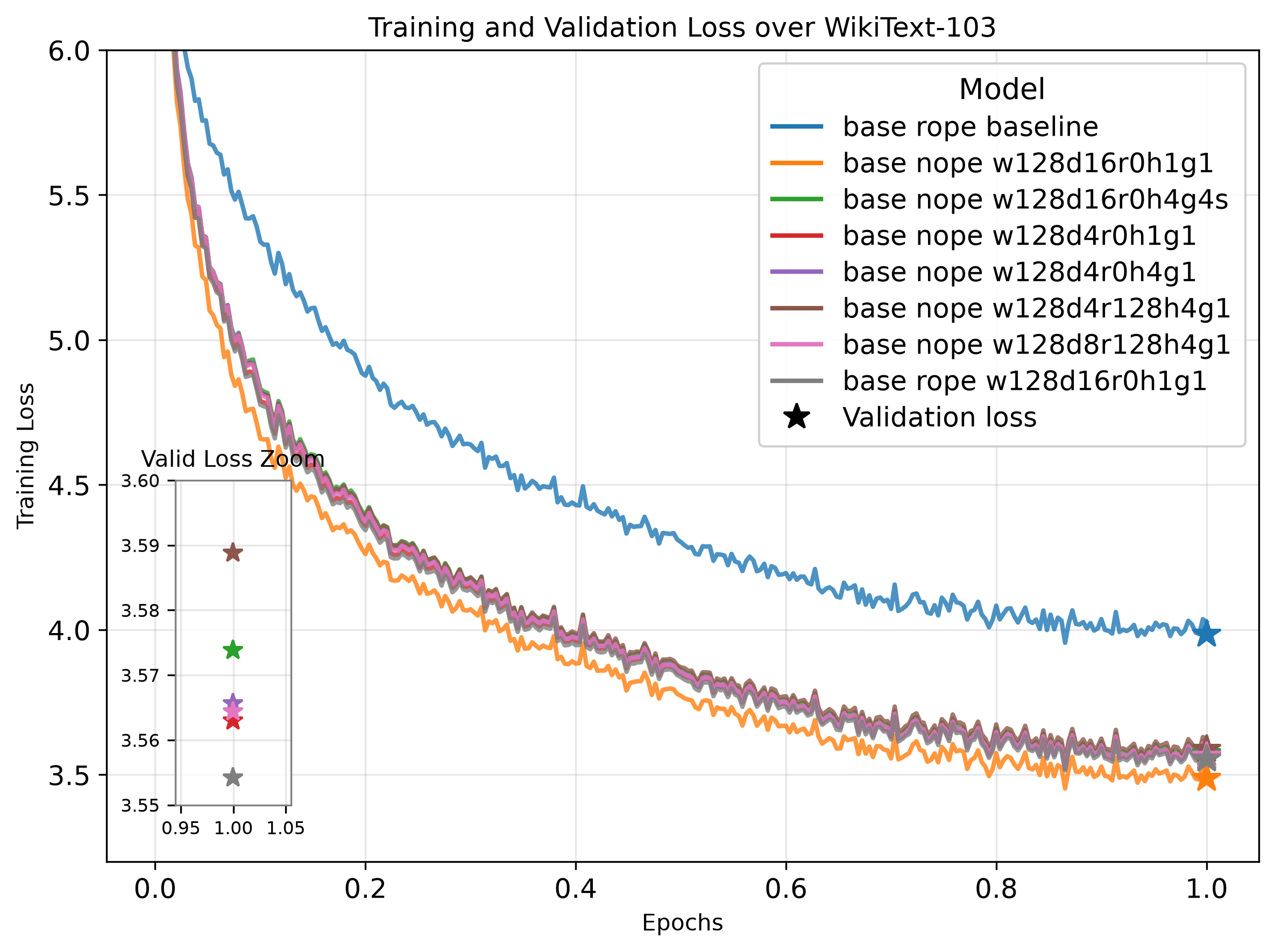}
        \caption{Training curves and validation loss of Base models on WikiText-103.}
        \label{fig:wt103b}
    \end{minipage}
\end{figure}

First, we train tiny models around \texttt{nope w128d6r0h3g1}. \Cref{fig:wt103t} shows that every ConvPE variant outperforms the RoPE baseline ($3.942$--$4.073$ vs.\ $4.249$). Increasing the projection dimension from $6$ to $12$ ($3.966\to3.942$) and the window from $64$ to $128$ ($4.047\to3.966$) give the largest gains. Grouped convolution hurts the most ($4.048$ for \texttt{g3} and $4.073$ for \texttt{g3s}), and a smaller window costs about as much ($4.047$). The rank-128 variant ($3.955$), RoPE integration ($3.962$), and the full kernel ($3.966$) are close, whereas $H_b=1$ is $0.011$ worse in this sweep ($3.977$).

We then train base models around \texttt{nope w128d4r0h4g1} and \texttt{nope w128d16r0h1g1}. \Cref{fig:wt103b} shows that every ConvPE variant again beats the RoPE baseline ($3.986$). The best configuration is \texttt{w128d16r0h1g1} ($3.489$). Because $H_b=1$ is also the cheapest bias-head setting, we adopt \texttt{w128d16r0h1g1} as the default.

We also tried feeding $Q$ (as in Composite~\cite{chang2021convolutionsselfattentionreinterpretingrelative}), $KV$, or $X$ directly to the convolution; all three performed worse than first projecting $X$ to the low-dimensional $X_{\mathrm{down}}$.

\subsection{Pretraining}
\label{sec:pretrain}

\begin{figure}[!ht]
    \centering
    \includegraphics[width=0.70\linewidth]{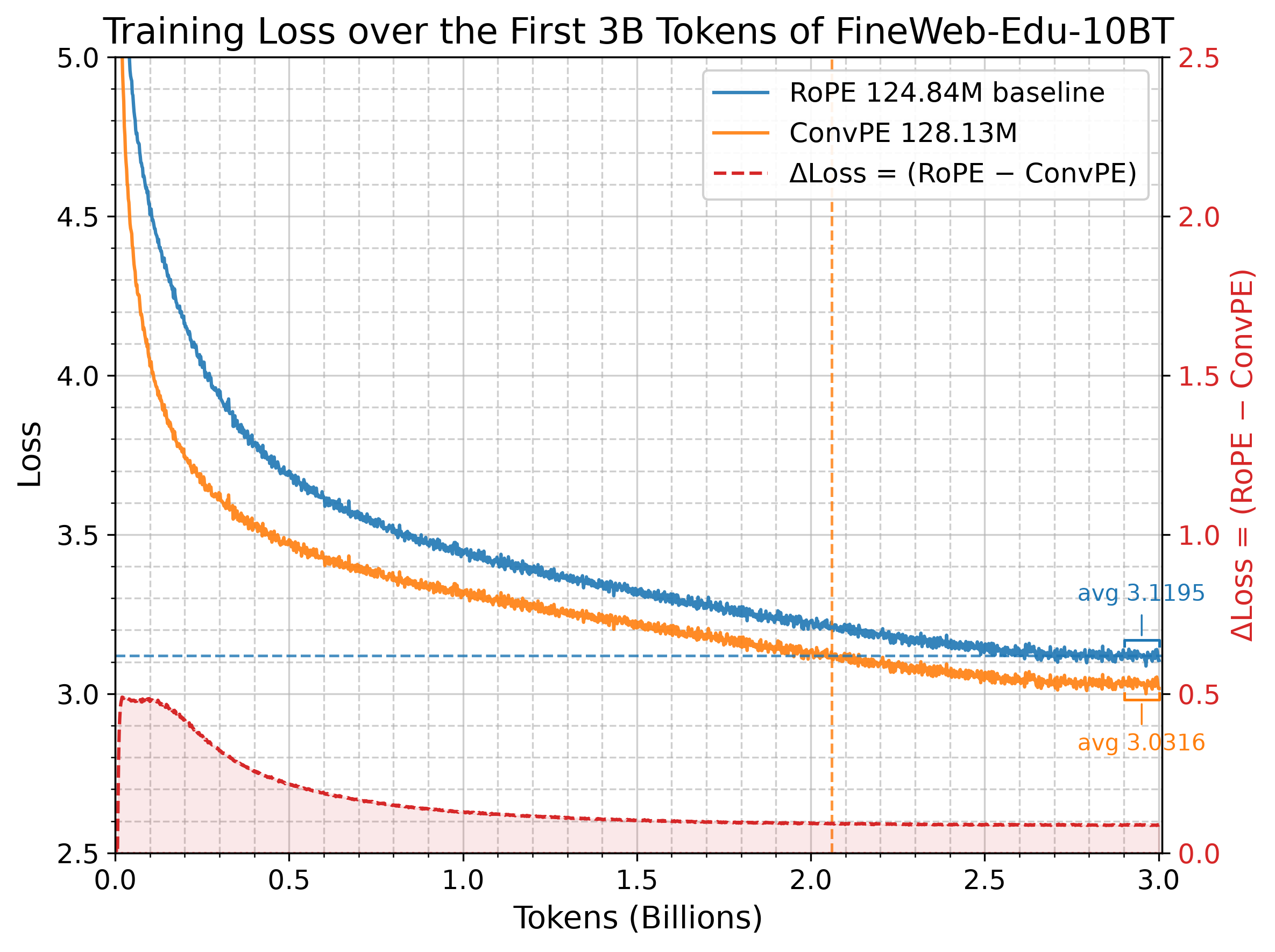}
    \caption{Training loss on the first 3B tokens of FineWeb-Edu-10BT for the RoPE baseline and ConvPE \texttt{w128d16}. The dashed curve is the loss difference (RoPE minus ConvPE).}
    \label{fig:pretrain}
\end{figure}

We train the RoPE baseline and ConvPE \texttt{w128d16} from scratch on the first 3B tokens of FineWeb-Edu-10BT with the recipe in \Cref{sec:setup}. \Cref{fig:pretrain} shows that ConvPE converges faster and reaches a lower final loss. Averaged over the last 0.1B tokens, the final training loss drops from $3.1195$ to $3.0316$ ($-2.82\%$ loss, $-8.41\%$ perplexity). ConvPE reaches the baseline's final loss after $2.06$B tokens instead of $3$B ($-31.33\%$).

To measure cost, we implement high-performance Flash MQA kernels in TileLang and benchmark the two configurations on one GPU with batch size $4$ and sequence length $1024$, averaged over $1{,}000$ iterations after $1{,}000$ warm-up iterations. ConvPE has $128.13$M parameters versus $124.84$M ($+2.64\%$), and its forward pass takes $23.14$ ms versus $22.53$ ms ($+2.71\%$). During training, ConvPE processes $2.60$ iterations/s versus $2.77$ ($6.5\%$ more time per step). The per-step overhead is larger than the forward-pass overhead, which is consistent with the ConvPE backward kernel being less optimized than its forward kernel. Combining the $31.3\%$ token reduction with the $6.5\%$ per-step overhead gives a net reduction of about $27\%$ in the wall-clock time needed to match the baseline's loss. The exact value depends on how the curves are smoothed, so we report it as approximate.

\subsection{Attention Analysis}
\label{sec:attn_analysis}

\begin{figure}[!ht]
    \centering
    \includegraphics[width=\linewidth]{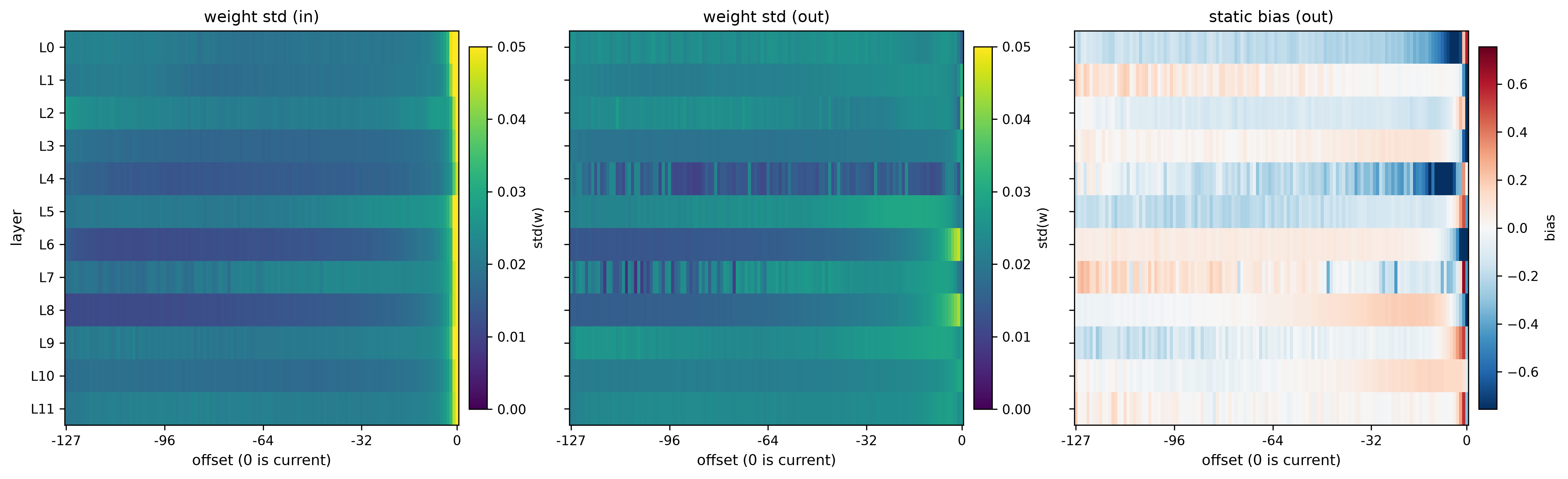}
    \caption{ConvPE statistics of the pretrained \texttt{base w128d16} model across layers. Left: standard deviation of the kernel weights in the input dimension. Middle: standard deviation in the output dimension. Right: static output bias, averaged over bias heads. The x-axis is the relative offset ($-127$ to $0$; $0$ is the current token).}
    \label{fig:attn}
\end{figure}

To understand the role of the ConvPE module, we analyze the kernel weights and the static output bias of the trained \texttt{base w128d16} model. Since the average weights are close to zero, the standard deviation indicates \textbf{feature selectivity}: a higher standard deviation means that the kernel focuses on specific channels. \Cref{fig:attn} shows three properties.

\textbf{1. Near-end dominance.}
Near the current token, both the static bias and the input-feature selectivity reach their largest magnitudes. This indicates that the positional relationship between nearby tokens is important: ConvPE applies highly specific feature extraction and strong positional priors to the immediate local context.

\textbf{2. Far-end relaxation.}
Toward the far end of the window, the standard deviation is lower and the static bias is close to zero for most layers. ConvPE therefore does not impose a strong static prior at this range, leaving the decision to the convolution over the dynamic input and to the query--key semantic match.

\textbf{3. Self-suppression and self-enhancement.}
The static bias has a clear layer-wise pattern at the current position. Layer 0 shows self-enhancement with a bias of $+0.5885$, whereas every deeper layer except L4 ($+0.0115$) shows self-suppression with a negative bias. This is consistent with deeper layers shifting attention away from the current token and toward the historical context.

\subsection{Downstream Evaluations}
\label{sec:eval}

\begin{table}[!ht]
\centering
\small
\begin{tabular}{llccc}
\toprule
Dataset & Metric & RoPE & ConvPE & Random \\
\midrule
WikiText-2 & word PPL $\downarrow$ & 46.56 & \textbf{42.05} & -- \\
LAMBADA~\cite{paperno2016lambada} & PPL $\downarrow$ & 118.10 & \textbf{88.49} & -- \\
LAMBADA & acc $\uparrow$ & 0.2507 & \textbf{0.2604} & -- \\
HellaSwag~\cite{zellers2019hellaswag} & acc & 0.2894 & \textbf{0.2949} & 0.25 \\
HellaSwag & acc-norm & 0.3148 & \textbf{0.3272} & 0.25 \\
XStoryCloze~\cite{lin2022xstorycloze} & acc & 0.5460 & \textbf{0.5526} & 0.5 \\
PIQA~\cite{bisk2020piqa} & acc & 0.6251 & \textbf{0.6338} & 0.5 \\
PIQA & acc-norm & 0.6208 & \textbf{0.6235} & 0.5 \\
ARC-Easy~\cite{clark2018arc} & acc & 0.5114 & \textbf{0.5421} & 0.25 \\
ARC-Easy & acc-norm & 0.4487 & \textbf{0.4701} & 0.25 \\
WinoGrande~\cite{sakaguchi2019winogrande} & acc & 0.5091 & \textbf{0.5193} & 0.5 \\
BoolQ~\cite{clark2019boolq} & acc & 0.5676 & \textbf{0.5893} & 0.5 \\
TruthfulQA-MC1~\cite{lin2021truthfulqa} & acc & \textbf{0.2399} & 0.2375 & 0.25 \\
\midrule
Average$^\dagger$ & acc & 0.4372 & \textbf{0.4475} & -- \\
\bottomrule
\end{tabular}
\caption{Zero-shot evaluation of the pretrained models on nine language tasks, using lm-evaluation-harness~\cite{gao2023lmeval}. Lower PPL and higher accuracy are better; the Random column gives the chance level where one exists. $^\dagger$Mean of one accuracy per task (acc-norm for HellaSwag, PIQA, and ARC-Easy, and acc otherwise), excluding the two perplexities.}
\label{tab:pe}
\end{table}

This section evaluates the pretrained ConvPE and RoPE models on downstream benchmarks without any fine-tuning. \Cref{tab:pe} shows that ConvPE improves every accuracy metric except TruthfulQA-MC1. The mean of the eight accuracy metrics rises from $43.72\%$ to $44.75\%$, a $2.35\%$ relative improvement. TruthfulQA-MC1 is the exception: both models fall below the $25\%$ random-choice baseline ($23.99\%$ and $23.75\%$), so this task does not separate them. Overall, the zero-shot results support ConvPE as a stronger position embedding than RoPE for the evaluated language tasks.

\subsection{Context Extension}

\begin{table}[!ht]
\centering
\small
\begin{tabular}{lcccccc}
\toprule
Context & RoPE & ConvPE & RoPE-8K & YaRN & ConvPE-8K & YaRN-ConvPE \\
\midrule
1K & 22.5272 & \textbf{20.672} (-8.2\%) & -- & -- & -- & -- \\
2K & 23.4439 & \textbf{20.1441} (-14.1\%) & -- & -- & -- & -- \\
4K & 67.8413 & \textbf{20.6415} (-69.6\%) & -- & -- & -- & -- \\
8K & 260.4544 & \textbf{27.8015} (-89.3\%) & 23.0702 & 21.6276 & \textbf{19.7405} (-8.7\%) & 21.3539 (-1.3\%) \\
\bottomrule
\end{tabular}
\caption{Validation perplexity for context lengths from 1K to 8K.}
\label{tab:valppl}
\end{table}

We first measure language-modeling perplexity on the validation dataset at increasing context lengths. Both models are initially trained with 1K-token contexts. \Cref{tab:valppl} shows a clear advantage for ConvPE as the context length grows. The perplexity of RoPE rises sharply beyond 2K, whereas ConvPE remains stable up to 4K. 

We further train four models: RoPE-8K, ConvPE-8K, YaRN and YaRN-ConvPE, starting from the pretrained models on an additional 150M tokens from FineWeb-Edu-10BT using an 8K context length. When evaluated at an 8K context length, ConvPE-8K achieves the lowest perplexity among the models. These results demonstrate the effectiveness of ConvPE for context extension.

We then evaluate these models on RULER~\cite{hsieh2024ruler}, which measures long-context retrieval and reasoning on controlled synthetic tasks. \Cref{tab:ruler} shows that ConvPE outperforms RoPE and YaRN on most tasks.

\begin{table*}[!ht]
\centering
\small
\setlength{\tabcolsep}{4pt}
\begin{tabular}{lcccccccccccc}
\toprule
& \multicolumn{2}{c}{1K} & \multicolumn{2}{c}{2K} & \multicolumn{5}{c}{4K} & \multicolumn{3}{c}{8K}\\
\cmidrule(lr){2-3}\cmidrule(lr){4-5}\cmidrule(lr){6-10}\cmidrule(lr){11-13}
Task & RoPE & Conv & RoPE & Conv & RoPE & Conv & Ro8K & YaRN &Co8K & YaRN &Co8K & YaRN-ConvPE \\
\midrule
niah\_single\_1   & 0.614 & \textbf{0.890} & 0.004 & \textbf{0.722} & 0.000 & \textbf{0.314} & 0.030 & 0.170 & \textbf{0.688} & 0.018 & \textbf{0.640} & 0.040 \\
niah\_single\_2   & 0.612 & \textbf{0.664} & 0.000 & \textbf{0.038} & 0.000 & 0.000 & 0.010 & 0.022 & 0.022 & 0.046 & 0.024 & \textbf{0.052} \\
niah\_single\_3   & 0.090 & \textbf{0.180} & 0.000 & \textbf{0.014} & 0.000 & 0.000 & 0.002 & 0.002 & \textbf{0.010} & 0.008 & \textbf{0.012} & 0.004 \\
niah\_multikey\_1 & 0.198 & \textbf{0.200} & 0.118 & \textbf{0.194} & 0.000 & \textbf{0.002} & 0.036 & 0.074 & \textbf{0.186} & 0.072 & \textbf{0.154} & 0.094 \\
niah\_multikey\_2 & \textbf{0.002} & 0.000 & 0.000 & 0.000 & -- & -- & -- & -- & -- & -- & -- & -- \\
niah\_multikey\_3 & 0.000 & \textbf{0.008} & 0.000 & 0.000 & -- & -- & -- & -- & -- & -- & -- & -- \\
niah\_multiquery & 0.104 & \textbf{0.186} & 0.029 & \textbf{0.179} & 0.000 & 0.000 & 0.016 & 0.016 & \textbf{0.089} & 0.021 & \textbf{0.026} & 0.022 \\
niah\_multivalue & 0.149 & \textbf{0.187} & 0.127 & \textbf{0.186} & 0.000 & 0.000 & 0.025 & 0.102 & \textbf{0.128} & \textbf{0.038} & 0.032 & 0.036 \\
ruler\_vt        & \textbf{0.032} & 0.008 & 0.000 & \textbf{0.052} & 0.000 & \textbf{0.107} & 0.000 & 0.000 & \textbf{0.015} & 0.000 & \textbf{0.025} & 0.000 \\
ruler\_cwe       & 0.002 & 0.002 & 0.000 & \textbf{0.002} & -- & -- & -- & -- & -- & -- & -- & -- \\
ruler\_fwe       & 0.001 & \textbf{0.012} & 0.000 & \textbf{0.001} & -- & -- & -- & -- & -- & -- & -- & -- \\
ruler\_qa\_squad & 0.148 & \textbf{0.191} & 0.078 & \textbf{0.091} & 0.017 & \textbf{0.052} & 0.064 & \textbf{0.093} & 0.075 & 0.049 & 0.044 & \textbf{0.053} \\
\midrule
Average$^\dagger$ & 0.163 & \textbf{0.211} & 0.030 & \textbf{0.123} & 0.001 & \textbf{0.040} & 0.015 & 0.040 & \textbf{0.101} & 0.021 & \textbf{0.080} & 0.025\\ 
\bottomrule
\end{tabular}
\caption{Accuracy on twelve RULER tasks~\cite{hsieh2024ruler}. Conv denotes ConvPE; Ro8K denotes RoPE-8K; Co8K denotes ConvPE-8K. Cells marked -- were not evaluated because both models were near zero on these tasks at the shorter lengths. $^\dagger$Average over all twelve tasks, with cells marked -- counted as zero.}
\label{tab:ruler}
\end{table*}

\subsection{2D Position Embedding}
\label{sec:cifar}

\begin{table}[!ht]
\centering
\small
\begin{tabular}{lccc}
\toprule
Model & Params (M) & CIFAR-10 & CIFAR-100 \\
\midrule
RoPE baseline (1D)    & 5.346 & 0.910 & 0.627 \\
ConvPE2D \texttt{w3d4} & 5.359 & 0.931 & \textbf{0.695} \\
ConvPE2D \texttt{w3d8} & 5.372 & \textbf{0.933} & 0.683 \\
ConvPE2D \texttt{w5d4} & 5.386 & 0.929 & 0.682 \\
ConvNet (reference)    & 4.336 & 0.945 & 0.737 \\
\bottomrule
\end{tabular}
\caption{Test accuracy of ViT-Tiny on CIFAR-10/100.}
\label{tab:cifar}
\end{table}

For 2D inputs, ConvPE2D replaces the 1D causal convolution with a 2D convolution over an $H\times W$ patch grid: each patch receives a bias over its $w\times w$ neighborhood, and the same window is shared across the grid. We use ViT-Tiny models with multi-head attention (MHA) and the configuration $N_{\mathrm{layers}}=12$, $D_{\mathrm{model}}=192$, $N_{\mathrm{heads}}=3$, and $D_{\mathrm{head}}=64$, with $4\times4$ patches ($64$ tokens). Models are trained on CIFAR-10 and CIFAR-100~\cite{krizhevsky2009learning} for $100$ epochs with batch size $256$, using AdamW at $3\times10^{-4}$ for the remaining parameters and Muon at $3\times10^{-3}$ for the backbone 2D weights. At test time we average the logits of each image and its horizontal flip. The RoPE baseline applies 1D RoPE to the flattened patch sequence; the 2D axial RoPE baseline is considered in \Cref{sec:peer}.

\Cref{tab:cifar} shows that ConvPE2D \texttt{w3d4} improves CIFAR-10 and CIFAR-100 accuracy by $2.31\%$ and $10.85\%$ relative to the 1D RoPE baseline, and \texttt{w3d8} gives the best CIFAR-10 result ($0.933$). The convolutional reference reaches $0.945$ and $0.737$, so ConvPE2D narrows but does not close the gap between ViT-Tiny and a CNN at this scale. The influence of the 2D hyperparameters requires further study on larger models and datasets.

\subsection{Comparison with Peer Methods}
\label{sec:peer}

To situate ConvPE among recent positional mechanisms, we use an independent reproduction by Zhijian Qiao (Hong Kong University of Science and Technology), who trained seven methods with the same 12-layer, 768-wide backbone, on the same first 3B tokens of FineWeb-Edu-10BT, at a global batch size of $16$. Evaluation uses the first 1M tokens of the training dataset. This reproduction leaves the input embedding and the output head untied, yielding models with $163$--$167$M parameters, compared with $125$--$128$M in our runs. Absolute values therefore differ from those in \Cref{tab:pe,tab:valppl,tab:ruler}, so comparisons should be made within \Cref{tab:lm-summary}. Details are provided in the ConvPE repository (\texttt{pe\_comparison.md}).

\begin{table*}[!ht]
\centering
\small
\setlength{\tabcolsep}{3.2pt}
\begin{tabular}{lccccccccccc}
\toprule
& & & & \multicolumn{4}{c}{FineWeb-Edu 1M PPL $\downarrow$} & Eval Acc & \multicolumn{3}{c}{RULER Acc} \\
\cmidrule(lr){5-8}\cmidrule(lr){10-12}
Method & Params (M) & ms/step & Mem (GiB) & 1K & 2K & 4K & 8K & (\%) $\uparrow$ & 1K & 2K & 4K \\
\midrule
NoPE~\cite{kazemnejad2023impact}      & 163.470 & \textbf{141.30} & 10.77 & 24.34 & 173.71 & 1389.25 & 3618.99 & 42.78 & 19.31 & 0.15 & 0.00 \\
Composite~\cite{chang2021convolutionsselfattentionreinterpretingrelative} & 163.498 & 180.02 & 11.01 & 21.34 & 38.15 & 246.16 & 984.34 & 44.37 & 25.97 & 0.88 & 0.09 \\
RoPE~\cite{su2021roformer}      & 163.470 & 149.03 & 10.77 & 22.89 & 24.19 & 55.01 & 207.58 & 43.86 & 11.96 & 2.63 & 0.26 \\
ALiBi~\cite{press2021train}      & 163.470 & \underline{145.67} & 10.77 & 21.37 & 20.77 & 20.60 & 20.52 & 43.90 & 14.37 & 3.80 & 1.61 \\
CABLE~\cite{veisi2025cable}      & 163.691 & 188.05 & 13.17 & 20.71 & 20.01 & \underline{19.68} & \underline{19.52} & \underline{44.72} & 19.80 & 7.69 & 3.22 \\
DAPE V2~\cite{zheng2025dapev2}      & 163.512 & 273.71 & 20.25 & \textbf{19.47} & \textbf{18.71} & \textbf{18.31} & \textbf{18.09} & \textbf{44.74} & \underline{26.11} & \textbf{17.09} & \textbf{7.45} \\
ConvPE    & 166.764 & 156.00 & 11.21 & \underline{20.52} & \underline{19.95} & 20.67 & 29.33 & 44.66 & \textbf{26.58} & \underline{14.46} & \underline{3.49} \\
\bottomrule
\end{tabular}
\caption{Comparison with peer position mechanisms in the independent reproduction. PPL is measured on the first 1M pretraining tokens; Eval Acc is the mean of the eight accuracy metrics in \Cref{tab:pe} (acc-norm where available); RULER Acc is the mean over the twelve RULER tasks at each length. All rows share the same backbone, data, and evaluation pipeline.}
\label{tab:lm-summary}
\end{table*}

\Cref{tab:lm-summary} shows that ConvPE performs well on both the perplexity diagnostic and the accuracy metrics. At 1K and 2K it has the second-lowest PPL after DAPE V2. At 4K and 8K, DAPE V2, CABLE, and ALiBi extrapolate better on this diagnostic. On RULER, ConvPE has the best mean accuracy at 1K and is second at 2K and 4K. Its mean Eval Acc is within $0.08$ points of the best result. 

DAPE V2 convolves the full attention score map, so it is $1.84\times$ slower per step than RoPE ($273.71$ vs.\ $149.03$ ms) and uses $20.25$ GiB of training memory compared with $10.77$ GiB for RoPE. Because it requires materializing the full score map, it cannot use FlashAttention and incurs quadratic complexity in the sequence length. These costs make DAPE V2 unsuitable for long-context training.

ALiBi~\cite{press2021train} and CABLE~\cite{veisi2025cable} are cheaper, but both retain a distance-penalty form. This prior is reasonable for local language modeling, but it conflicts with long-context retrieval, where the relevant evidence may lie far from the query. Prior work supports this concern: Kazemnejad et al.~\cite{kazemnejad2023impact} find that ALiBi and other explicit position embedding methods are not well suited for length generalization on downstream tasks. RULER exhibits the same pattern in our reproduction: ConvPE outperforms both ALiBi and CABLE at every length.

\begin{table}[!ht]
\centering
\small
\setlength{\tabcolsep}{4pt}
\begin{tabular}{lccc}
\toprule
Method & Params (M) & ms/step & CIFAR-100 acc (\%) $\uparrow$ \\
\midrule
NoPE~\cite{kazemnejad2023impact}             & 5.363 & 66.45 & 56.03 \\
2D ALiBi~\cite{fuller2023croma}       & 5.363 & 66.63 & 61.38 \\
Composite2D~\cite{chang2021convolutionsselfattentionreinterpretingrelative}         & 5.596 & 78.29 & 63.50 \\
2D RoPE~\cite{heo2024ropevit}       & 5.363 & 78.44 & 64.17 \\
Learnable 2D abs. PE~\cite{dosovitskiy2021image} & 5.376 & \textbf{63.50} & 64.36 \\
Fixed 2D sin-cos~\cite{dosovitskiy2021image}    & 5.363 & 66.65 & 65.07 \\
CABLE2D~\cite{veisi2025cable}           & 5.377 & 98.90 & 65.25 \\
DAPE2D V2~\cite{zheng2025dapev2}            & 5.374 & 92.74 & 66.01 \\
ConvPE2D           & 5.377 & 82.03 & \textbf{67.98} \\
\bottomrule
\end{tabular}
\caption{CIFAR-100 test accuracy in the independent reproduction (ViT-Tiny, 100 epochs). Bold marks the best value per column; parameter counts are not ranked. ConvPE2D is the \texttt{w3d4} variant.}
\label{tab:cifar100}
\end{table}

\Cref{tab:cifar100} reports the CIFAR-100 comparison with 2D adaptations of the same methods. ConvPE2D outperforms all baselines. These results indicate that the convolutional bias transfers to grid-structured inputs, although the absolute ranking of 2D position mechanisms remains sensitive to the backbone and training recipe.

\section{Conclusions and Future Work}
\label{sec:conclusion}

ConvPE adds a learned, windowed bias to attention scores which is compatible with RoPE, attention masks, and KV concatenation, and it naturally supports multidimensional extensions. In our experiments it improves pretraining and validation perplexity, evaluation accuracy, context extension perplexity and accuracy, and CIFAR accuracy at a small parameter and computational overhead. Among the compared methods, ConvPE combines competitive downstream accuracy with a FlashAttention-compatible design; when accuracy, parameter count, per-step cost, and compatibility are considered jointly, it is the most favorable choice for long-context training.

Our study has two main limitations. First, the main experiments use a single seed and models of up to $167$M parameters, so the size of the gains at larger scale is unknown. Second, the max context length evaluated is 8K. We view these as directions for future work:

\begin{itemize}
    \item \textbf{Kernel optimization.} Further optimize the backward and decode kernels.
    \item \textbf{Scaling behavior.} Investigate ConvPE with larger Transformers and larger datasets.
    \item \textbf{Long-context modeling.} Evaluate ConvPE at larger contexts.
    \item \textbf{Architectural integration.} Combine ConvPE with Sliding Window Attention (SWA)~\cite{beltagy2020longformer}, Compressed Sparse Attention (CSA)~\cite{deepseek2026v4}, and other Transformer variants.
    \item \textbf{More modalities.} Extend ConvPE to video, audio, multimodal inputs, and control models.
\end{itemize}

\section*{Acknowledgements}

We thank Zhijian Qiao (Hong Kong University of Science and Technology) for the independent reproduction and the peer comparison in \Cref{sec:peer}.

\bibliographystyle{unsrtnat}
\bibliography{references}

\end{document}